\documentclass[journal,twoside,web]{ieeecolor}
\usepackage{tmi}
\usepackage{cite}
\usepackage{amsmath,amssymb,amsfonts}
\usepackage[bb=boondox]{mathalfa}
\usepackage{mathtools}
\usepackage{algorithmic}
\usepackage{graphicx}
\usepackage{stfloats}
\usepackage{textcomp}
\usepackage{arydshln}
\usepackage[dvipsnames]{xcolor}
\usepackage{hyperref}
\hypersetup{
  colorlinks   = true,
  urlcolor     = blue,
  linkcolor    = blue,
  citecolor   = blue 
}
\newcommand*{\red}{\textcolor{red}}
\newcommand*{\orange}{\textcolor{Dandelion}}

\newcommand*{\green}{\textcolor{ForestGreen}}
\definecolor{accessblue}{cmyk}{.24,1,1,0}

\usepackage{pifont}
\newcommand{\cmark}{\ding{51}}%
\newcommand{\xmark}{\ding{55}}%
\begin{document}
\title{

Federated Learning with Research Prototypes for Multi-Center MRI-based Detection of Prostate Cancer with Diverse Histopathology
}
\author{Abhejit~Rajagopal, Ekaterina~Redekop, Anil~Kemisetti, Rushi~Kulkarni, Steven~Raman, Kirti~Magudia, Corey~W.~Arnold~(IEEE Member), Peder~E.~Z.~Larson~(IEEE Member)
\thanks{A.~Rajagopal, A.~Kemisetti, and P.E.Z.~Larson are with the Department
of Radiology and Biomedical Imaging, University of California, San Francisco, 94158 USA. E.~Redekop, R.~Kulkarni, S.~Raman, and C.~Arnold are with the Department of Radiology, Univeristy of California, Los Angeles, 90024 USA. K.~Magudia is with the Department of Radiology, Duke University, Durham, Durham 27705 USA. Correspondence e-mail: abhejit.rajagopal@ucsf.edu.
Funding support of NIH/NIBIB (\#F32EB030411), NIH/NCI (\#R01CA229354 and \#R21CA220352).}}
\markboth{Submitted to IEEE Transactions on Medical Imaging, 2022}
{Rajagopal \MakeLowercase{\textit{et al.}}: Federated Learning with Research Prototypes for MRI-based Detection of Prostate Cancer}
\maketitle
\begin{abstract}
Early prostate cancer detection and staging from MRI are extremely challenging tasks for both radiologists and deep learning algorithms, but the potential to learn from large and diverse datasets remains a promising avenue to increase their generalization capability both within- and across clinics. To enable this for prototype-stage algorithms, where the majority of existing research remains, in this paper we introduce a flexible federated learning framework for cross-site training, validation, and evaluation of deep prostate cancer detection algorithms. Our approach utilizes an abstracted representation of the model architecture and data, which allows unpolished prototype deep learning models to be trained without modification using the NVFlare federated learning framework. Our results show increases in prostate cancer detection and classification accuracy using a specialized neural network model and diverse prostate biopsy data collected at two University of California research hospitals, demonstrating the efficacy of our approach in adapting to different datasets and improving MR-biomarker discovery. We open-source our \texttt{FLtools} system, which can be easily adapted to other deep learning projects for medical imaging.
\end{abstract}
\IEEEpeerreviewmaketitle
\section{Introduction}
\IEEEPARstart{P}{rostate} cancer is the most prevalent cancer in American men but data shows that it affords a 99\% survival rate if the cancer is detected early \cite{sung2021global}. Compared with the low sensitivity of PSA blood tests and the potential complications of invasive biopsy, screening based on magnetic resonance imaging (MRI) offers the potential of a fast, safe~(non-invasive), and localized detection of prostate cancer that can aid in both diagnosis and treatment planning ~\cite{turkbey2012multiparametric}. Unfortunately, the current PIRADSv2 radiological standard for reading MRI has been shown to correlate poorly with the Gleason grade (cancer severity) determined by biopsy and histopathological analysis, with a positive predictive value of just 35\%~\cite{epstein20162014, westphalen2020variability}. As such, there is great interest in developing data-driven machine learning algorithms to assist and improve radiologists' capabilities in detecting and staging of clinically-significant prostate cancer (CS-PCa).

However, to-date accurate and early MRI-based detection of prostate cancer has eluded even deep learning algorithms, with most results in the literature either focused on staging of severe disease (Gleason patterns $\geq 3+3$)~\cite{cao2019joint}
or achieving similar accuracy to PIRADS using more-balanced screening populations~\cite{schelb2019classification}.
A persistent issue in these works is validating and improving the generalization behavior (i.e.~accuracy) of models, both within institutions and across institutions, where MRI protocols, granularity of biopsy data, MRI hardware, and patient populations can vary significantly~\cite{mehralivand2022deep,sarma2021federated}. 
This is especially important for screening populations where prostate cancer is typically less salient in MRI, and single-site training may suffer from inadvertent bias or brittleness that ultimately limits performance even on data collected at the same institution.

\begin{figure}[t!]
    \centering
    \includegraphics[width=\columnwidth]{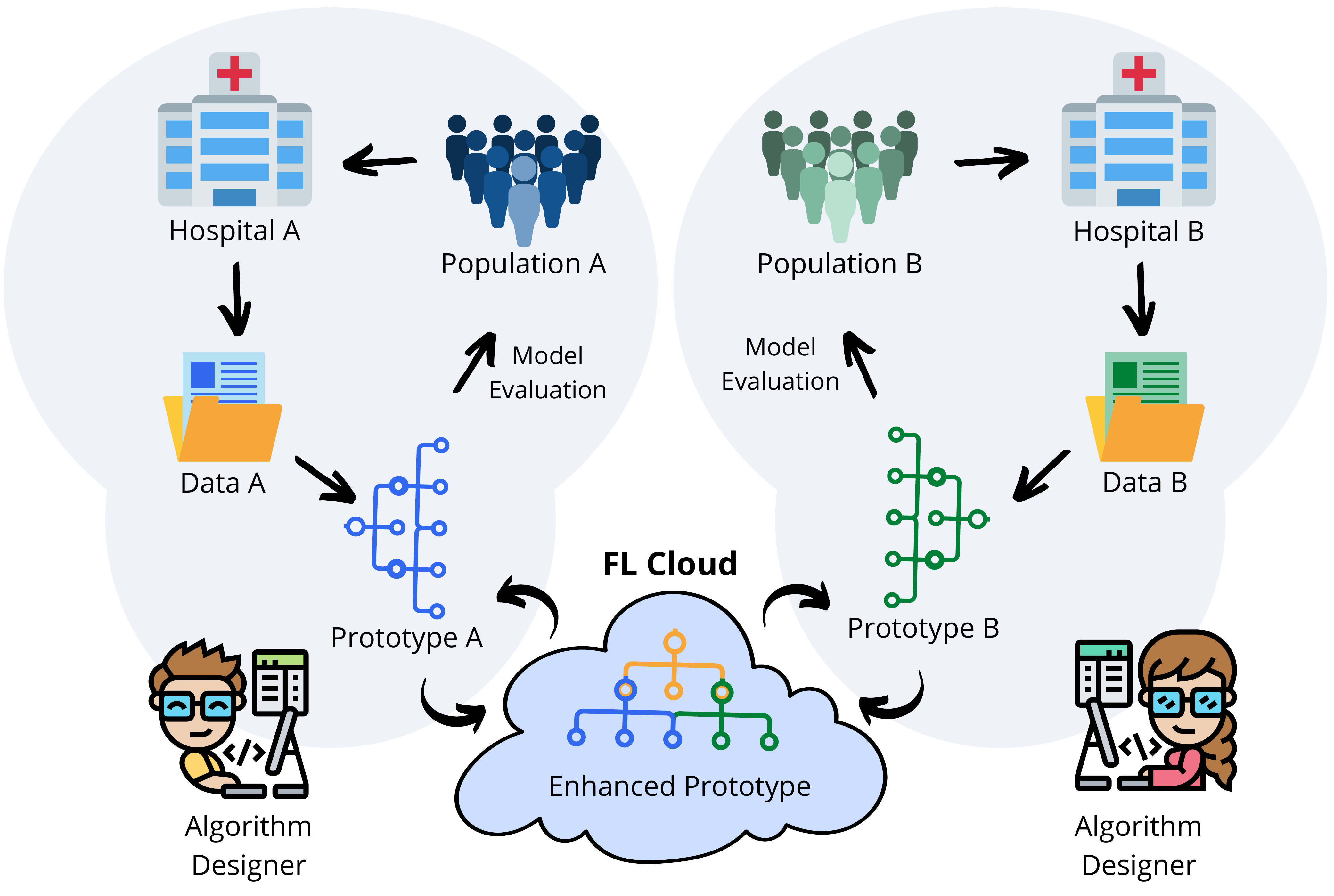}
    \caption{Federated learning of research prototype algorithms enables cross-site validation that is useful for improving within-site performance.}
    \label{fig:federated1}
\end{figure}

Federated learning (FL) is a presents an opportunity here to overcome these barriers, by offering a platform to share models and abundant data for extensive cross-site validation and enhanced model training~(Fig.~\ref{fig:federated1}) without the need to share images or other sensitive protected health information~(PHI). However, existing FL implementations have many requirements, such as the use of only-standard neural network architectures, freezing of model evaluation code and data pipelines, or restrictions on the format of the data, which creates barriers for inclusion of research sites and specialized approaches~\cite{antunes2022federated}. This is especially prohibitive for medical imaging problems where patient examination protocols differ by research site, deep learning is still exploratory, and the community has not agreed on an ideal model architecture or data format, as is the case for the assessment of prostate cancer with MRI.

To this end, we introduce a framework for model- and data-agnostic federated learning of research prototype algorithms that enables cross-site validation and joint-training of deep neural networks (DNNs) with \textit{zero} modifications to existing site-specific model architecture code or local training routines. We achieve this vision by defining data and model abstractions (currently implemented for \texttt{pytorch}) that are used jointly by local- and federated-training routines, but can crucially differ across research sites. Our open-source toolkit, \texttt{FLtools}, is built on top of the NVidia \texttt{NVFlare} framework \cite{nvflare}, and also includes a simulation system (\texttt{FLsim}) that allows for model debuging and inspection of intermediate tensors while applying the same routines used in the full-scale FL system.

We demonstrate the efficacy of this system on the cross-institution training of a custom deep learning model for multi-parametric MRI-based detection and classification of prostate cancer.  For this we use our proposed "UCNet", which utilizes a fully-convolutional backbone, but features a 3D region-of-interest (ROI) classification head that enables per-lesion, per-sextant, and ultimately per-exam prediction of prostate cancer severity. The UCNet model achieves this by utilizing a histogram representation of the International Society of Urological Patholog (ISUP) grade group that reflects a common data representation for prostate cancer histopathology across research sites, which may themselves include exams and groundtruth pathology data at various granularities.

Other medical imaging applications that would directly benefit from this approach include classification of less common cancers, e.g.~in the brain or in the kidneys, where the ability to utilize data from multiple sites and all sources of histology groundtruth jointly can significantly increase dataset sizes and the potentially boost overall detection accuracy. The presented FL design pattern is especially useful for data-starved deep learning problems that are challenging enough to necessitate the use of non-standard architectures or loss functions, where its desirable to expand dataset size and diversity while retaining model, dataloader, and learning algorithm mutability.

\subsection{Prior Work}

Prior work in federated learning includes both problems of theoretical interest (learning strategies under various constraints, such as communication bottlenecks, non-iid data, or privacy concerns) and of practical value; with the latter focusing on the development of tools and frameworks to alleviate various user constraints. \cite{mcmahan2021advances} provides an excellent overview of various available frameworks and research problems. One of the primary practical challenges an FL system faces is making the workflow as straightforward as possible, ideally approaching the ease-of-use achieved by ML systems for local (single computer) training. In this paper we address this by developing a library and design pattern that specifically extends NVidia's NVFlare toolkit for federated training.

Specifically for prostate cancer detection from MRI, several works have utilized federated learning. \cite{sarma2021federated} multi-center federated training improves prostate gland segmentation, an important sub-step in the search for prostate cancer biomarkers. \cite{yan2020variation} proposed a variation-aware federated learning framework where where the variations among clients are minimized by transforming the images (ADC maps) of all clients onto a common image space. Both these works, however, utilize homogeneous annotation \textit{datatypes}, which is an important variation both within and across institutions.
Our work addresses this with a model architecture that leverages highly heterogeneous radiological annotation and histopathology. We realize this vision with multi-center federated learning (FL).

\subsection{Contributions}
Thus our contributions are as follows:
\begin{itemize}
    \item We introduce a model and data abstraction for prostate MRI cancer detection and classification algorithms that supports intra- and inter-site exam heterogeneity.
    \item We introduce a modular design pattern for federated learning (FL) that uses these abstractions to enable synchronous cross-site training with \textit{zero} modifications to existing research code and minimal engineering overhead.
    \item We demonstrate the utility of this approach by training a custom deep learning model (UCNet) that is well-suited to the format and uncertainties associated with prostate MRI and associated histopathology, using data from two University of California hospitals (1800+ exams total).
    \item We open-source our code: \href{https://gitlab.com/abhe/prostate-mpmri/-/tree/FL}{gitlab.com/abhe/prostate-FL}
\end{itemize}

\begin{figure*}[b]
    \centering
    \includegraphics[width=\linewidth]{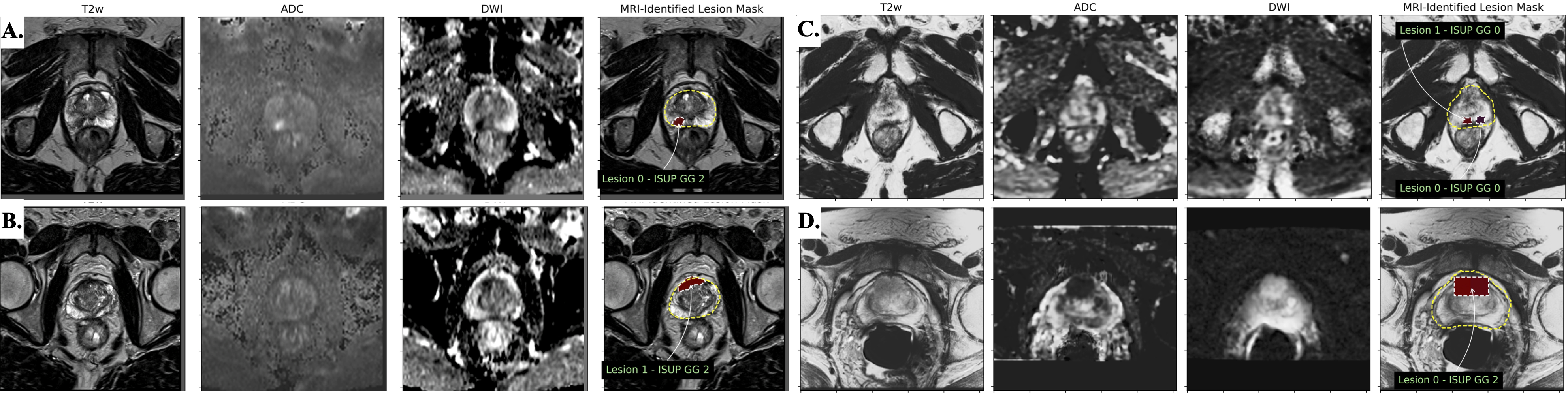}
    \caption{Intra- and inter-site variations of multiparametric MRI data. (A,B)~Natural variation between appearance of ISUP grade group 2 lesions (countoured) in UCLA data. (C,D)~Large variation in the apparent MR contrast and size of lesion annotations (bounding boxes) in UCSF data.}
    \label{fig:variations}
\end{figure*}
\section{Dataset and Assumptions}

\subsection{Multi-parametric Magnetic Resonance Imagery}
Multi-parametric magnetic resonance imagery (mp-MRI) prostate exams are composed of T2-weighted imaging, diffusion-weighted imaging (DWI), and dynamic contrast-enhanced (DCE) MRI. T2-weighted MRI provides structural information based on water content and tissue composition, DWI and associated apparent diffusion coefficient (ADC) maps represent the restriction of water movement in tissue that is typically altered due to the increased density of prostate tumors compared to healthy tissue, and DCE provides measurements of vasculature that is altered by tumors. Thus, mp-MRI provides a rich combination of information, but may vary in appearance due to subtle differences in choice of MR pulse sequence, scanner hardware, and physiologic variability in healthy prostate and tumors~(Fig.~\ref{fig:variations}). In this paper, the T2-weighted and DWI as well as the associated ADC maps from mp-MRI series were reconstructed and subsequently registered (downsampled) to the same spatial resolution of [0.66, 0.66, 2.24]~mm for x, y, and z axis respectively for all series in datasets from both research sites. 

\subsection{Histopathology Derived from Prostate Biopsy}
Prostate MRIs are typically conducted after an indication of a high prostate-specific antigen (PSA) blood concentration, or age. At both research sites, a board-certified radiologist annotated the mp-MRI series with possible lesions, suggesting areas for prostate biopsy. At UCSF, the prostate biopsy is conducted by transrectal-ultrasound (TRUS) MR-guided biopsy, where a T2 series is fused with the ultrasound to help navigate the needle and target the MR-annotated regions. In addition to targeted biopsy, at UCSF urologists systematically sample the prostate in 6 regions, providing additional confidence to the gland-wise cancer designation. Unfortunately, as the prostate is highly non-rigid, only coarse coordinates are associated with each biopsy core sample. That is, for UCSF data we assume the location of the systematic biopsies based on a geometric division of the prostate into sextants in the registered MR-image space, and we assume the targeted biopsy occurs in a bounding prism around each MR-identified lesion. The groundtruth histopathology for each biopsy sample was determined by a pathologist who observed stained slices of the biopsy core under a microscope and assigned a Gleason pattern to each. We convert this Gleason pattern to the standardized ISUP grade group (1-5, 0 for negative), where a grade group $\geq 2$ indicates clinically significant prostate cancer (CS-PCa).

At UCLA, a similar targeted TRUS prostate biopsy is conducted, but with additional innovation to identify the location of needle in the joint ultrasound-MRI image space, and enhanced radiologist-defined contouring of each lesion. The histopathology for each biopsy core is determined in a similar fashion as UCSF, resulting in a set of ISUP grade group scores. In this paper however, for data from the UCLA site we assume the highest ISUP grade group in each exam is known, but not to which lesion they correspond. For the UCLA site, we also do not include any systematic biopsy data. This mismatch in available data provides additional consideration for the design and supervision of our chosen deep architecture and supervision system, as will become apparent in Section~\ref{sec:architecture}.

\subsection{Gland Segmentation and Contrast Normalization}
As T2 and DWI have arbitrary non-quantitative image amplitudes, we apply interquartile range (IQR)-based intra-image normalization to address the relative nature of MR image intensity values both within and across research sites and to eliminate outlying values created by imaging artifacts. Specifically, each image was normalized to the image-level IQR computed within the 3D prostate gland (annotated by a radiologist or previously developed neural network segmentation model \cite{sarma2021harnessing}) according to~\cite{pellicer2022deep}:
\begin{align}
    I_{norm} = \frac{I- \text{percentile}(I, 1)}{\text{percentile}(I, 99) - \text{percentile}(I, 1)}
\end{align}

Z-score image normalization followed IQR-based normalization to overcome the problem of high variability of intensity distribution between different patients by transforming the intensities to have zero mean and unit variance.

\subsection{Summary of Datasets at Research Sites}
Table~\ref{tab:datasets} summarizes the distribution of ISUP grade groups in training, validation, and testing exams for both research sites. As evident, our data is highly non-iid with respect to the cancer grade distribution, in addition to subtle differences in the characteristics of the imagery.
\begin{table}[hbt!]
    \centering 
    \resizebox{1.0\linewidth}{!} {
    \begin{tabular}{c|c|c|c|c|c|c|}
        \textbf{} & UCSF-Train & UCSF-Val & UCSF-Test & UCLA-Train & UCLA-Val & UCLA-Test \\
        \hline
        max ISUP 0   & 92  & 17 & 24 & 196 & 26 & 43 \\
        max ISUP 1   & 222 & 27 & 73 & 172 & 24 & 31 \\
        max ISUP 2   & 228 & 30 & 61 & 197 & 27 & 40 \\
        max ISUP 3-5 & 137 & 22 & 40 & 172 & 24 & 34 \\
        \hline
        Totals       & 679 & 96 & 198 & 737 & 101 & 148
    \end{tabular}
    }
    \caption{Number of exams as a function of ISUP grade group.}
    \label{tab:datasets}
\end{table}

\begin{figure*}
    \begin{minipage}[b]{0.38\linewidth}
        \centering
        \includegraphics[width=\linewidth]{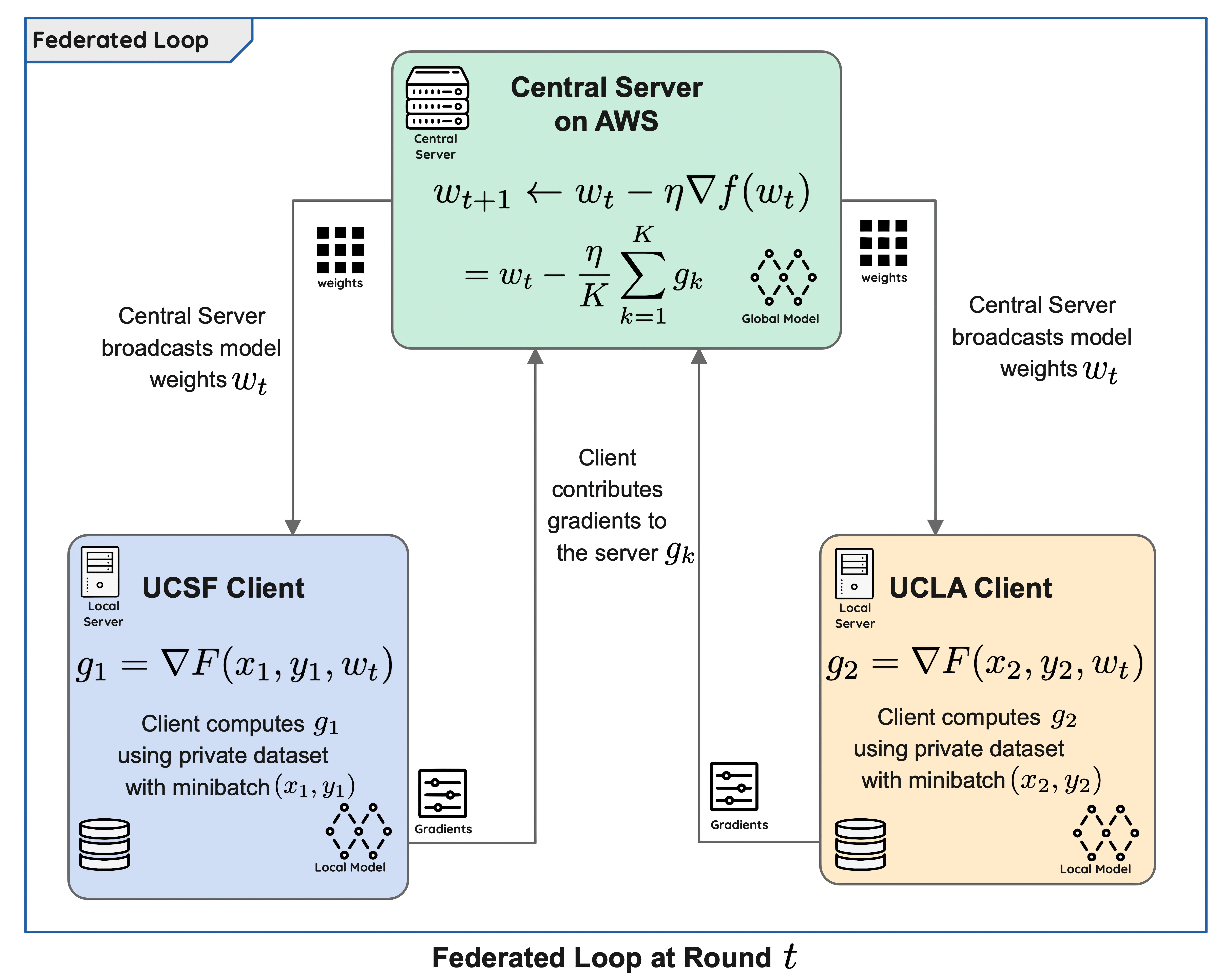}
        \caption{FL topology for \texttt{FedSGD}}
        \label{fig:federated_model}
    \end{minipage}
    \hfill
    \begin{minipage}[b]{0.61\linewidth}
    \resizebox{\linewidth}{!} {
    \begingroup
    \renewcommand{\arraystretch}{1.2}
    \begin{tabular}{l|c|c|c|c}
         & \multicolumn{1}{c|}{   NVFlare 1.0} 
         & \multicolumn{1}{c|}{   NVFlare 1.0} 
         & \multicolumn{1}{c|}{   NVFlare 2.0} 
         & \multicolumn{1}{c}{    Our System} \\
         
         & \multicolumn{1}{c|}{+ Clara 4.0} 
         & \multicolumn{1}{c|}{+ BYOT} 
        & \multicolumn{1}{c|}{Custom} 
         & \multicolumn{1}{c}{$($~NVFlare} \\
         
         & \multicolumn{1}{c|}{} 
         & \multicolumn{1}{c|}{+ BYOC} 
         & \multicolumn{1}{c|}{FLComponents} 
         & \multicolumn{1}{c}{ + \texttt{FLTools}$)$} \\
         
        \hline
        Use the Clara Models for Training & \cmark &\cmark & \cmark & \cmark \\
        Bring your model without code modification & \xmark & \xmark & \xmark & \cmark \\ 
        Use custom dataloader & \xmark & \cmark & \cmark & \cmark \\
        Check FL loop simulation without NVFlare API & \xmark & \xmark & \xmark & \cmark \\ 
        Fix training issues using python debugger & \xmark & \xmark & \cmark (poc mode) & \cmark \\ 
        Custom Checkpointing and Transfer Learning & \xmark & \cmark & \cmark & \cmark \\ 
        Using the local logging without modification & \xmark & \xmark & \xmark & \cmark \\
    \end{tabular}
    \endgroup
    }
    \caption{Comparison of Features in Different Federated Learning Frameworks.}
    \label{tab:FL}
    \end{minipage}
\end{figure*}

\section{Modular Framework for Federated Learning}
In this paper, we are using federated learning (FL) as a way to combine and maximize the use of non-sensitive (anonymized) prostate MRI exams across research sites with minimal impact to research prototype algorithm design and existing validation pipelines. 

In this context, FL involves adapting a ``local'' training loop into a ``federated'' training loop, where training is synchronized between the clients and a central server to implement learning algorithms like federated stochastic gradient descent (\texttt{FedSGD}) or weight averaging (\texttt{FedAverage}). However, this synchronization demands advanced engineering and researchers consequently often rely on FL frameworks to implement the details and help them build federated loops quicker. To achieve this, many frameworks (e.g.~NVidia Clara) use programming design patterns like dependency injection to maintain control of the main program logic (e.g.~looping structures) while also providing programming hooks for custom code. Although this simplifies the interface and increases adoption for well-calibrated production-ready models in established problem domains, this presents an obstacle for researchers developing new models in problem domains at their infancy (like prostate cancer detection from MRI). This is because the researcher cannot easily develop, debug, and refactor the core model or training logic as a module independent of the FL framework engineering code.

To address this, we present a design pattern for FL that separates model development and FL implementation code, providing a rich FL development environment medical imaging \textit{researchers}. Our approach utilizes Nvidia’s NVFlare, an independent python library from Nvidia that allows researchers to collaboratively train deep learning models without sharing patient data~\cite{nvflare}. In addition to ``Bring Your Own Trainer'' (BYOT) capabilities, NVFlare includes a ``Bring Your Own Component'' (BYOC) mode that enables users to develop their own FL components, e.g.~to extract, aggregate, serialize, and transfer model weights and gradients between clients, using high-level APIs separated from the deep learning model architecture and data-specific code, such as pre-processing or model execution. We develop these components as a lightweight library of modules, \texttt{FLtools}, that can be imported into existing research repositories, enabling a separation between local model development and FL deployment. Furthermore, \texttt{FLtools} includes a simulation tool \texttt{FLsim} that enables researchers to test FL deployment on one or more systems outside of NVidia's federated environment (Table~\ref{tab:FL}).

\subsection{System Design and Components}
To separate FL deployment from model development, we define a deep learning design pattern composed of 4 elements:

\subsubsection{Model Abstraction} We adopt the \texttt{pytorch lightning}~\cite{falcon2019pytorch} design pattern for specifying models, which involves defining a class with the key methods:
\begin{itemize}
    \item \texttt{forward} -- a function that implements a forward pass or inference given a batch of \textit{input} data, using the model architecture and weights (defined elsewhere in the class), resulting in a set of tensors as the model's \textit{output}.
    \item \texttt{training\_step} -- a function that accepts a batch of training data (paired \textit{input} data and \textit{groundtruth} data), calls \texttt{forward} with \textit{input} data to produce \textit{output} data, and computes \textit{metrics} and \textit{losses} by comparison with the \textit{groundtruth} to monitor and improve model performance.
    \item \texttt{validation\_step} -- (optional) a function that performs the same function on \texttt{training\_step} using a frozen version of the model on a batch of validation data.
    \item \texttt{unpack\_batch} -- (optional) a function we define (extending \texttt{pytorch lightning}) to accept and unpack a dictionary of collated training or validation data, to enable easy dispatch of \texttt{forward} or \texttt{training\_step} calls.
    \item \texttt{configure\_optimizers} -- a function that returns a neural network optimization function of the user's choice.
\end{itemize}

\subsubsection{Data Abstraction} There are two abstractions we use here, one for the general FL design pattern, and the second for the specific MRI-based prostate cancer detection problem.

For the general FL design pattern, we require each client to provide a \texttt{get\_objects} function that returns the following:
\begin{itemize}
    \item \texttt{model} -- a python class, defined with the aforementioned member functions and learnable algorithm parameters.
    \item \texttt{train\_dataloader} -- a python iterable that (when iterated on) returns a collated dictionary of batched training data using client-specific code and data sources.
    \item \texttt{val\_dataloader} -- same as \texttt{train\_dataloader}, but returning a collated dictionary of validation data.

\end{itemize}
Thus, from the perspective of the FL training loops, data yielded by client-specific dataloaders are fed directly to \texttt{training\_step}, without additional specification or connector code within the FL environment. This provides flexibility for prototype algorithm developers at each research site to change model and data specifications (e.g~DNN toplogy, file loading, pre-processing, augmentation) without modifying the interface. Moreover, each site is free to modify and optimize local implementations of all the aforementioned components (including choice of training objectives and metrics), albeit within constrains of the federated algorithms employed.

Specifically, for the MRI-based prostate cancer detection problem, we define the content of the training data using a flexible dictionary interface. Besides imaging data (provided as a registered, multi-channel 3D tensor, $x\in\mathbb{R}^{3 \times X \times Y \times Z}$), we also include 3D binary region masks $y\in\mathbb{R}^{R \times X \times Y \times Z}$ representing each of the $R$ regions where histopathology data is included (lesions and sextants). Crucially, for this work we encode the histopathology data using a $z \in \mathbb{Z}^{R \times 2}$ matrix, representing a \textit{supervision signal} \{0,1,2\} and a maximum \textit{ISUP grade group} (0-negative, 1-5) for each region. The supervision signal is used to dynamically select learning objectives applicable to the type of histopathology groundtruth on a region-by-region basis, enabling large heterogeneity in the types of exams that can be included for training both within and across prostate cancer research sites. This operation is clarified in Section~\ref{sec:architecture}.

\subsubsection{Federated Toolkit (\texttt{FLtools})} 
\begin{figure}[t]
    \centering
    \includegraphics[width=\linewidth]{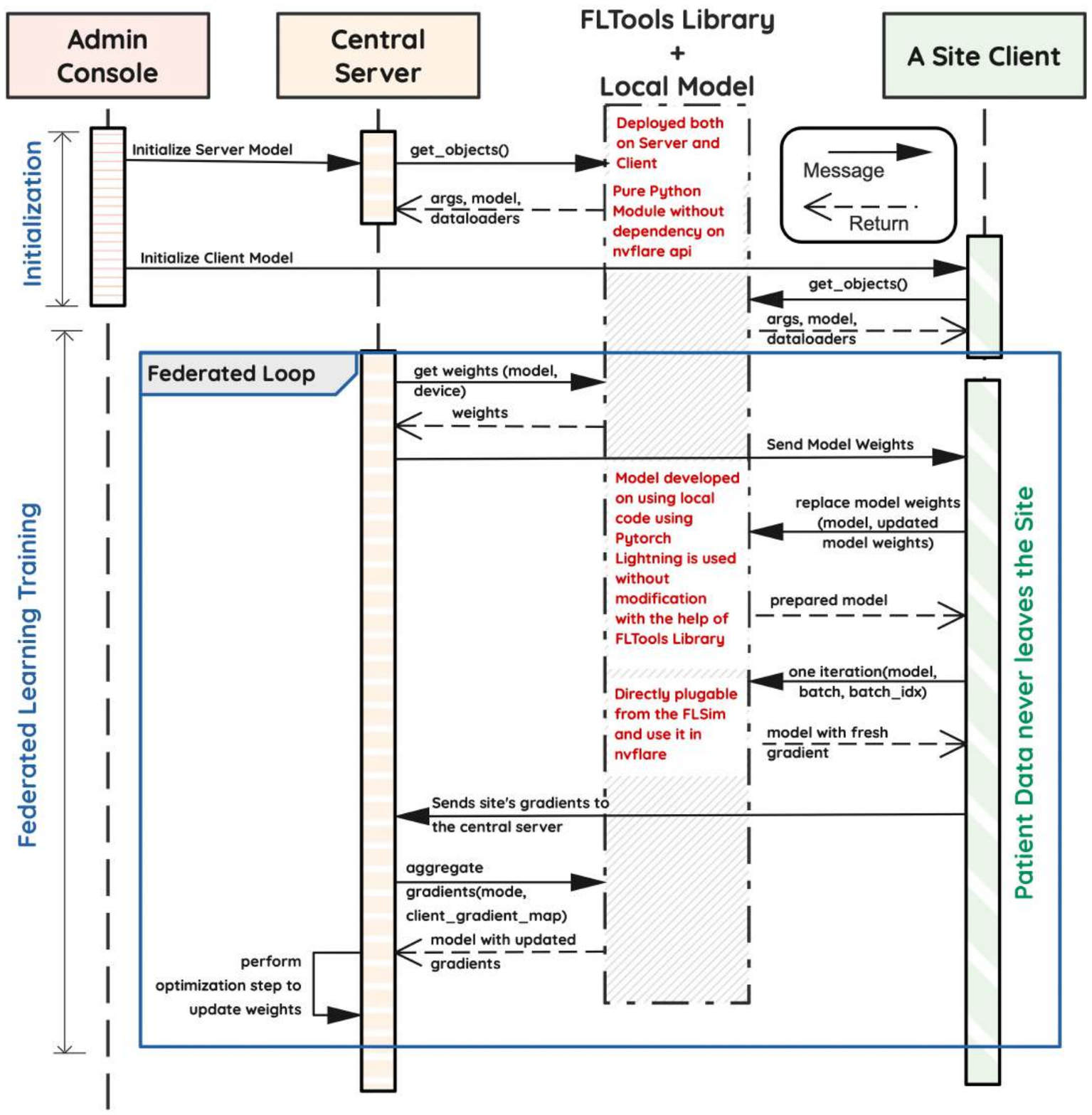}
    \vspace{-4mm}
    \caption{Modular federated system architecture}
    \label{fig:sequence1}
\end{figure}

Our FLtools library includes several components that represent baseline implementations of various NVFlare's components, but which are crucially reusable across models and FL experiments:
\begin{itemize}
    \item \texttt{FLComponents} - BYOT \& BYOC implementations utilizing the get\_objects interface. This decouples the model implementation from NVFlare's library, enabling reuse across models and experiments.
    \item \texttt{FLTrainer} - A reusable module extending the ``Trainer'' component, which uses the \texttt{get\_objects} interface to perform the training on the clients.
    \item \texttt{FLAggregator} - A reusable module that extends the ``Aggregator'' component, responsible for (a) validating the gradient contribution from each clients, and (b) aggregating  gradient averaging using \texttt{FLUtils}.
    \item \texttt{FLSharableGenerator} - A reusable module that extends the ``Sharable Generator'' component and converts a model to a sharable, a dictionary of weights and gradients that can be transmitted between the server and clients.
    \item \texttt{FLModelPersistor} - A reusable model that extends "ModelPersistor", responsible for saving and loading of the checkpoint file on the server.
\end{itemize}
Figure~\ref{fig:sequence1} depicts a sequence diagram elucidating the communication between various entities present in the federated topology implemented for \texttt{FedSGD}~(Fig.~\ref{fig:federated_model}). An important feature of \texttt{FLtools} is the clear separation between the engineer’s duties responsible for the machine learning operations (MLOps) team and the researcher developing the model. We found that the autonomy of the respective experts to work in separate areas and integrate via a programming contract fosters an environment of collaboration, conducive to FL success.

\begin{figure}[hbt!]
    \centering
    \includegraphics[width=0.73\linewidth]{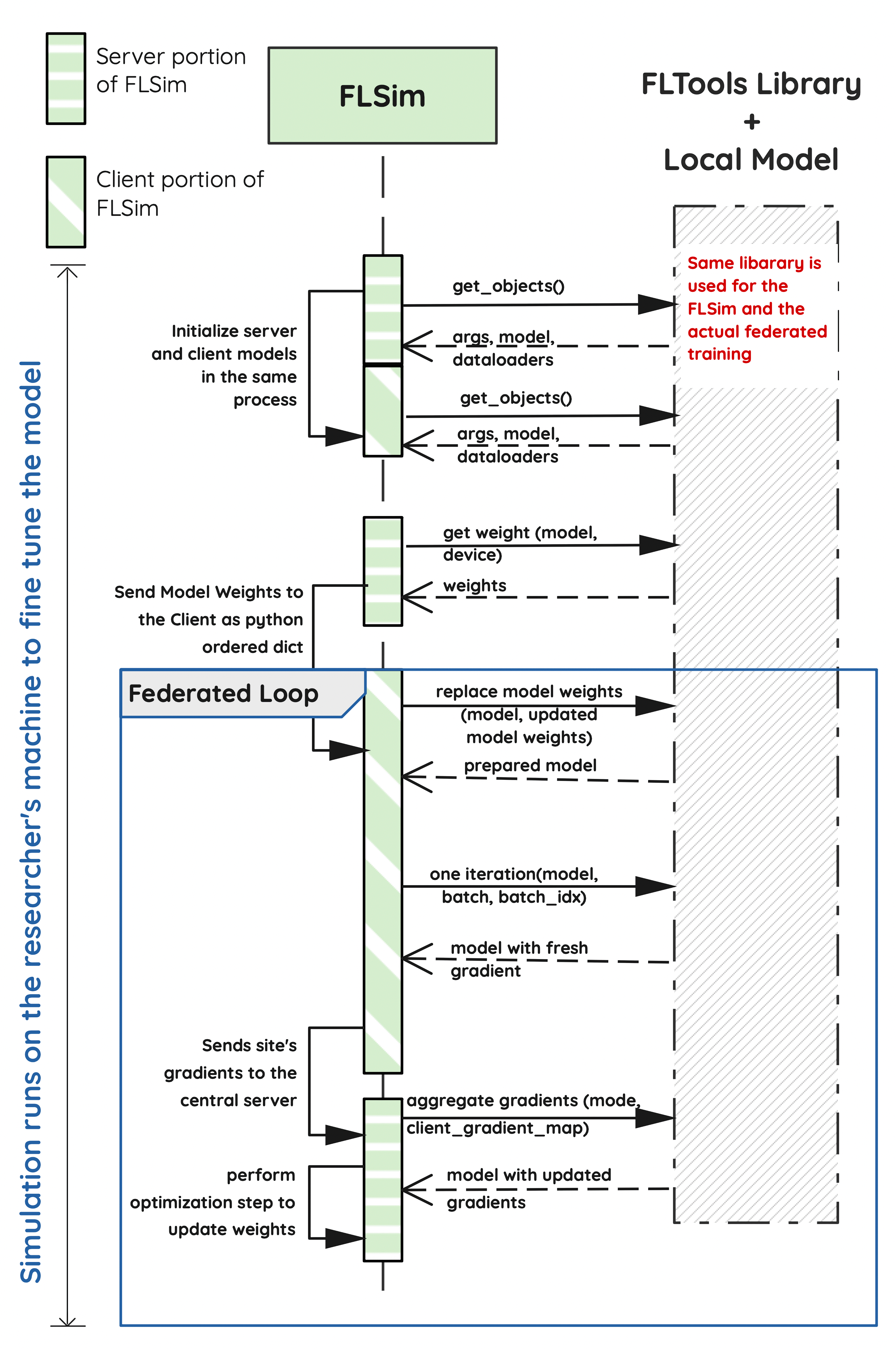}
    \vspace{-2mm}
    \caption{FLSim Local Federated Simulation}
    \label{fig:sequence2}
    \vspace{-4mm}
\end{figure}
\subsubsection{Local Simulation (\texttt{FLsim})}
Figure~\ref{fig:sequence2} shows how \texttt{FLsim} simulates the federated loop (Fig.~\ref{fig:sequence1}) on the local machine of the researcher. Since \texttt{FLSim} is decoupled from NVflare, the researcher is free to use existing debug tools (e.g.~\texttt{pdb}) to fine-tune model or computational hyperparameters~(e.g.~batch size) prior to running the federated training via NVFlare.

\subsection{Model Training and Implementation Details}
For the FL experiments in this paper, we host the central server on an Amazon Web Services (AWS) EC2 instance, and two clients that are behind institutional firewalls at UCSF and UCLA, respectively~(Fig.~\ref{fig:federated_model}). The central server aggregates gradients from each client and performs a weight update with appropriate momentum terms~(we are using the AdamW optimizer with \texttt{FedSGD}). Our FL concept involves private data, but also private metrics, so each client has no knowledge of how well the federated model is performing at other research sites. Instead, each institution's client monitors performance on the institution's own private validation dataset. Thus, there is no stopping criteria from the perspective of the server, and client can freely choose to select whichever checkpoint they consider to be most performant.

\begin{figure*}
    \centering
    \includegraphics[width=\linewidth]{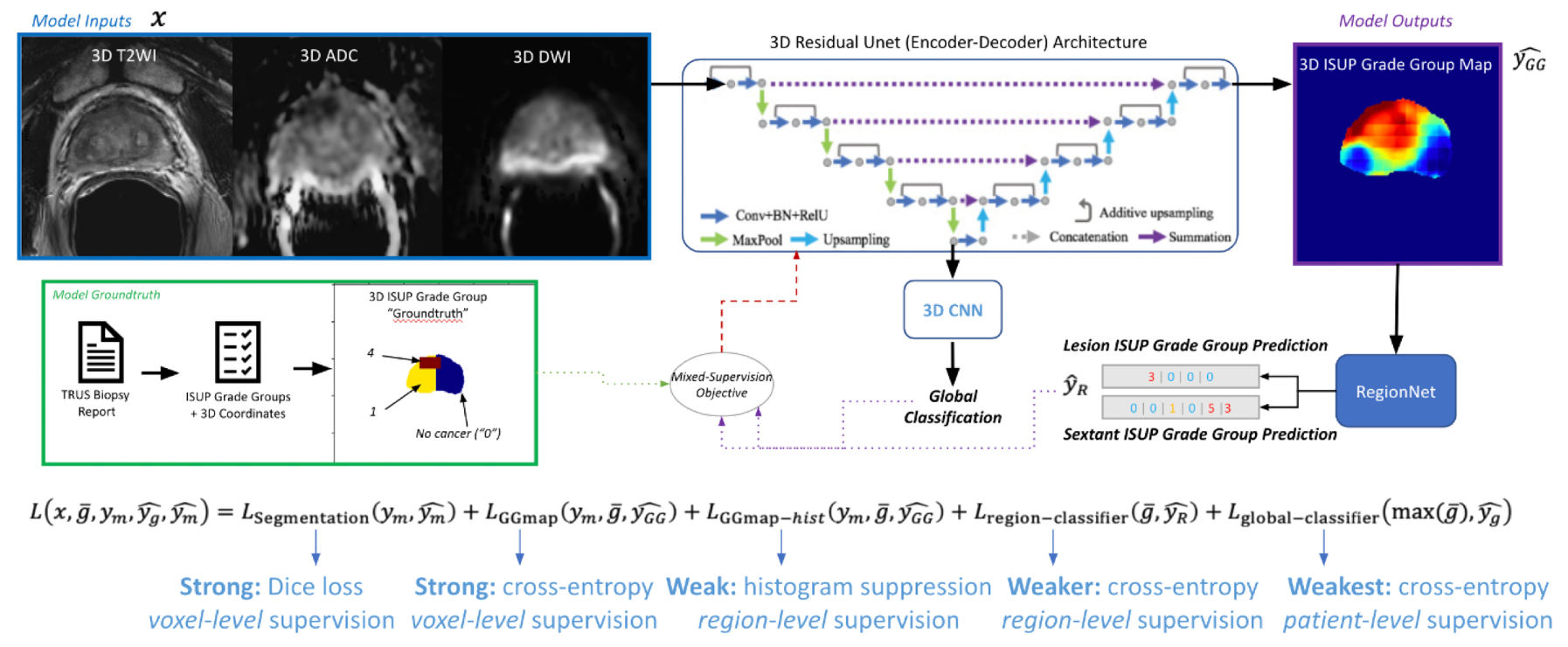}
    \caption{UCNet Architecture, depicted here with a 3D residual UNet backbone, histopathology-based histogram suppression, and regional classification modules. In this paper, UCNet takes registered 3D mp-MRI as input and produces as output: lesion segmentation maps, 1-hot-encoded cancer grading maps (for classification of clinically-significant prostate cancer, $K=2$), and per-region classifications ($\mathcal{L}_\text{global}$ not trained). }
    \label{fig:UCNet}
\end{figure*}

\section{UCNet for Mixed Histopathology Supervision} \label{sec:architecture}
For this problem, we utilize a 3D residual UNet with an additional fully-connected classification output head, an architecture we call ``UCNet''. Our UCNet implementation takes 3D mp-MRI as input and predicts 3D lesion segmentation maps, 3D ISUP grade group maps, and region-wise histograms that are used to determine region-wise and exam-wise cancer severity. We believe this architecture is representative of many deep learning tasks in medical imaging, but is particularly suited to handling the variety of groundtruth data available for prostate cancer. Of particular importance here is the dynamically-populated multi-task objective we use to train UCNet using highly heterogenous data within and across institutions.

\subsection{Spatial Prediction}
\subsubsection{Backbone} We use a fully-convolutional 3D residual UNet \cite{ronneberger2015u} as the backbone for UCNet, which in this case accepts a image tensor $x\in\mathbb{R}^{3 \times X \times Y \times Z}$ as input, and produces voxel-wise tanh-activated lesion maps $\hat{y}_\text{seg}\in\mathbb{R}^{1 \times X \times Y \times Z}$ and softmax-activated grade group membership predictions $\hat{y}_\text{gg}\in\mathbb{R}^{K \times X \times Y \times Z}$, where $K\in\mathbb{Z}_{[0,5]}$ represents the number of unique grade groups chosen for regression. For example, $K=2$ in the case of binary detection of CS-PCa (ISUP grade groups $\geq$ 2). $K$ is an adjustable hyperparmeter, but we find $K=2$ yields slightly higher validation accuracy for detection of CS-PCa compared to multi-class grading approaches.

\subsubsection{Strong Supervision of Lesion Segmentation and Grading}
We use a combination of dice and weighted binary cross-entropy losses to supervise the radiological lesion segmentation output, $\hat{y}_\text{seg}$, given the groundtruth segmentations $y_\text{seg}$, since this has been shown to achieve higher accuracy~\cite{rajagopal2021understanding}:
\begin{align}
    \mathcal{L}_\text{segmentation}(y_\text{seg}, \hat{y}_\text{seg}) = \mathcal{L}_\text{seg-dice} + \mathcal{L}_\text{seg-BCE}
\end{align}
\begin{align}
    \mathcal{L}_\text{seg-dice}(y_\text{seg}, \hat{y}_\text{seg}) = 1 - \frac{y_\text{seg} \cdot \hat{y}_\text{seg}}{y_\text{seg} + \hat{y}_\text{seg} + \epsilon}
\end{align}
\begin{align}
    \mathcal{L}_\text{seg-BCE}(y_\text{seg}, \hat{y}_\text{seg}) =
    \sum_z
    \sum_{x, y_\text{seg}=z}
    \frac{
    \text{BCE}(y_\text{seg}(x), \hat{y}_\text{seg}(x))
    }{|z| \cdot n_z}
\end{align}
where $\epsilon$ is chosen arbitrarily small to prevent overflow, $\text{BCE}$ represents the binary cross-entropy function, $n_z$ represents the number of voxels in class $z\in\{0,1\}$ and $|z|$ is the number of classes. Here, $y_\text{seg}$ is determined from as the voxel-wise max of the region masks $y\in\mathbb{R}^{R \times X \times Y \times Z}$, over regions where the supervision signal in the first column of the histopathology matrix $z\in\mathbb{R}^{R \times 2}$ is 1 to represent MRI-annotated lesions.

Similarly, we apply categorical binary cross entropy loss to each voxel of the predicted grade membership maps, $\hat{y}_\text{gg}\in\mathbb{R}^{K \times X \times Y \times Z}$, given one-hot encoded groundtruth histopathology data $y_\text{gg} \in \mathbb{R}^{R \times K}$ from regions $R^+$ where $z$ holds a supervision signal of 1 and a regression grade group that is not NaN. This is an important subtlety that allows supervision of segmentation outputs from the UCLA dataset even when the a grade group is not known for individual lesions. This strong voxel-wise grading or classification loss is then defined as:
\begin{align}
    \mathcal{L}_\text{GGmap}(y, z, y_\text{gg}, \hat{y}_\text{gg}) = \frac{1}{|R^+|} &\sum_{r \in R^+}  \sum_{k=1}^K y_\text{gg}[r,k] \log{\hat{y}_\text{gg}[\mathbb{1}_r]_k} \nonumber
\end{align}

where $\mathbb{1}_r$ represents an indicator (Kronecker delta) function selecting the voxels corresponding spatially to region $r$.

\subsection{Regional Prediction}
Unfortunately, strong voxel-wise supervision signals are not available for all histopathology data types, such as biopsy data. This is because the biopsy core only targets one, small portion of the prostate gland, in either a lesion or a regular tissue. For targeted lesion biopsies, we make the assumption that the biopsy sample is representative of a \textit{homogeneous} cancer profile within a tumor that is reasonably localized in the MRI coordinate system, enabling use of the aforementioned strong classification objective, $\mathcal{L}_\text{strong}$. Systematic biopsies, however, are neither localized nor representative of the cancer profile in \textit{heterogeneous} tissue throughout the gland, preventing the use of such strong voxel-wise objectives.

To this end, we utilize two weak supervision objectives in regions $R^*$ where the supervision signal in $z$ is $\geq 1$:
\subsubsection{Histogram Suppression}
Rather than regressing by value with a voxel-wise classification loss, we regress \textit{by distribution} by computing and penalizing a histogram of predicted voxel-wise grade groups for each region of interest. This can be achieved naturally (maintaining model differentiability) by selecting $K$-dimensional voxels of $\hat{y}_\text{gg}$ using the corresponding region masks in $y$, and computing their $K$-dimensional average, resulting in a set of approximate histograms $\hat{h}\in\mathbb{R}^{R^* \times K}$.

For regions $R^\alpha$ where the supervision signal is 1 (individual lesions), we assume the biopsy core represents a homogeneous cancer profile, so we penalize non-zero histogram bins that don't correspond to the groundtruth grade group:
\begin{align}
    \mathcal{L}_\text{hist-strong}(z, y_\text{gg}, \hat{h}) = \frac{1}{|R^\alpha|} &\sum_{r \in R^\alpha}  \sum_{k=1}^K y_\text{gg}[r,k] \log{\hat{h}[r,k]} \nonumber
\end{align}

For regions $R^\beta$ where the supervision signal is $>1$ (sextants, or exams where only the highest grade is known), we assume the biopsy core represents the highest cancer in a heterogeneous tissue, so we only penalize non-zero histogram bins that represent grade groups \textit{higher} than the groundtruth regression grade group:
\begin{align}
    \mathcal{L}_\text{hist-high}(z, y_\text{gg}, \hat{h}) = \frac{1}{|R^\beta|} &\sum_{r \in R^\beta}  \quad \sum_{
        \mathclap{\substack{k \; > \\
        \underset{k}{\mathrm{argmax}} \; y_\text{gg}[r] 
        }}
    }^K
    y_\text{gg}[r,k] \log{\hat{h}[r,k]} \nonumber
\end{align}
The net effect of these losses $\mathcal{L}_\text{GGmap-hist} = \mathcal{L}_\text{hist-strong} + \mathcal{L}_\text{hist-high}$ is to suppress the proportion of voxels representing grade groups not supported by the histopathology data and its expected uncertainty. Note that when $\mathcal{L}_\text{hist-high}=0$, the  histogram bin corresponding to the groundtruth grade group may not have the highest proportion (as expected for heterogeneous tissue).

\subsubsection{Regional Grading} To provide a clear indication of cancer severity in such cases, without relying on manual interpretation of the histograms, we feed the regional histograms $\hat{h}$ into a small dense network with ReLU activation in the final layer. This results in multi-hot encoded vectors $\hat{z}^{R^* \times K}$, indicating the presence of various ISUP grade groups in regions $R^*$.

To maintain the clearest interpretation of $\hat{h}$, in this paper we use a single-layer ReLU-activated network with tunable bias and weight-matrix frozen as identity $I_K$. The bias, thereby representing a set of ``optimal'' thresholds for declaring clinically-significant cancer in given region, is optimized through backpropagation, just as other parameters in the fully-convolutional portion of UCNet, via the loss:
\begin{align}
    \mathcal{L}_\text{region-classifier}(z, y_\text{gg}, \hat{z}) = \frac{1}{|R^*|} &\sum_{r \in R^*}  \sum_{k=1}^K y_\text{gg}[r,k] \log{\hat{z}[r,k]} \nonumber
\end{align}

While $R^*$ generally includes both segmented sextant and lesion regions, its worth noting that this can differ from the total region count $R$, e.g.~in exams where individual lesion contours are available but the individual scores are not.

\subsection{Global Prediction}
The UCNet architecture supports global prediction modules, which can be based either on the spatial map that is rendered by the decoding branch or the fully-encoded features produced by the encoding branch, of the fully-convolutional backbone. We do not use either capability in this paper, and instead focus on lesion-wise cancer classification accuracy.

\subsection{Multi-task Objective}

Then, our multi-task objective is:
\begin{align}
    \mathcal{L}(x, y, y_\text{gg}, z) = 
    \alpha_1 \lambda_1 \mathcal{L}_\text{region-classifier}
    + \alpha_2 \lambda_2 \mathcal{L}_\text{GGmap-hist} \label{eq:objective} \\
    + \alpha_3 \lambda_3 \mathcal{L}_\text{GGmap}
    + \alpha_4 \lambda_4 \mathcal{L}_\text{segmentation}  \nonumber
\end{align}
where we empirically choose $\bar{\lambda}=[1.0, 0.5, 1.0, 1.0]$, and $\bar{\alpha}$ is chosen based on the availability of different annotation groundtruth available for each exam in a minibatch, as well as the the type of supervision signals desired in the experiment. In this paper, we choose $\bar{\alpha}=[1.0, 1.0, 1.0, 1.0]$ to use  all described objectives whenever possible. We use the AdamW optimizer, initialized with an initial learning rate of $0.0015$ both locally and during federated training, to optimize the parameters of UCNet with respect to Equation~\ref{eq:objective}.

\subsection{Inference}
Inference is performed in a similar fashion to training, except with all augmentation turned off. Note that, due to the nature of weak histopathology groundtruth, it is not straightforward to train UCNet using volumetric patches. Instead, we feed full (registered and resampled) multi-contrast prostate MRI exams as input to UCNet, producing the collection of predictions for each region over the whole prostate gland.

\section{Results}
The federated learning system that we designed was successful in enabling multi-center training of the UCNet model, while avoiding having to share and centralize institutional patient data. Tables~\ref{tab:lesion_classification}-\ref{tab:lesion_classification} below compares the test-set performance of various model checkpoints trained locally and using FL, selected at each site by finding the checkpoint with the highest validation performance \textit{at each site}. This is an important point that leads to two versions of the federated model, as each site may select different checkpoints as ``best''. 

\begin{table}[hbt!]
    \centering
    \begin{tabular}{l|c|c}
        Model & UCSF & UCLA \\
        \hline
        UCSF-local & 0.134 & 0.0 \\
        UCLA-local & 0.000 & 0.15 \\
        UCSF-FL & 0.120 & 0.1\\
        UCLA-FL & 0.111 & 0.06 \\
    \end{tabular}
    \caption{Lesion Segmentation Intersection-over-Union (IoU)}
    \label{tab:lesion_segmentation}
\end{table}
\begin{table}[hbt!]
    \centering
     \begin{tabular}{c|l|c|c}
        Model & UCSF & UCLA \\
        \hline
        UCSF-local & 68.0\% [0.74, 0.63] & 53.3\%  [0.85, 0.51]\\
        UCSF-FL & 67.9\% [0.76, 0.60] & 62.8\% [0.69, 0.57]\\
        \hline
             & \orange{$\downarrow$ 0.01\%} & \green{$\uparrow$ 9.5\%} \\
        \\
        UCLA-local & 47.9\% [0.93, 0.03] & 69.5\% [0.67, 0.73]\\
        UCLA-FL & 62.7\% [0.78, 0.47] & 66.8\% [0.74, 0.65] \\
        \hline
             & \green{$\uparrow$ 14.8\%} & \red{$\downarrow$ 2.7\%} \\
    \end{tabular}
    \caption{Region-wise Lesion Binary Classification Accuracy. Bracketted numbers indicate TNR and TPR, respectively.}
    \label{tab:lesion_classification}
    \vspace{-4mm}
\end{table}

From Table~\ref{tab:lesion_classification} we can see that both local models performed well on their own private test sets, but the performance on the set from the different institution was much lower. UCLA local model showed 47.9\% accuracy on UCSF data, which is 30\% lower than UCSF local model performance. The result of UCLA model evaluation on UCSF data showed high TNR~(0.93) and low TPR~(0.03), which means low usability of the model on data in another institution. UCSF local model showed slightly higher accuracy of~53.3\% on the UCLA data, which was still 23\% lower than UCLA local model result. Although the best FL model was chosen on each site separately based on the local validation performance,both FL models succeeded in generalization. UCLA-FL model increased the UCSF data classification accuracy by 14.8\% with subsequent increase in TNR and TPR, which made the model usable in practise by another institution. The same result achieved UCSF-FL model, which increased UCLA local data performance accuracy by 9.5\% with increase in TNR and TPR.

\begin{figure*}
    \textbf{A: Evaluation of UCNet on UCSF test data, where FL reclaims accuracy lost by cross-site evaluation (Row 2).}\\
    {
    \centering
    \includegraphics[width=\linewidth]{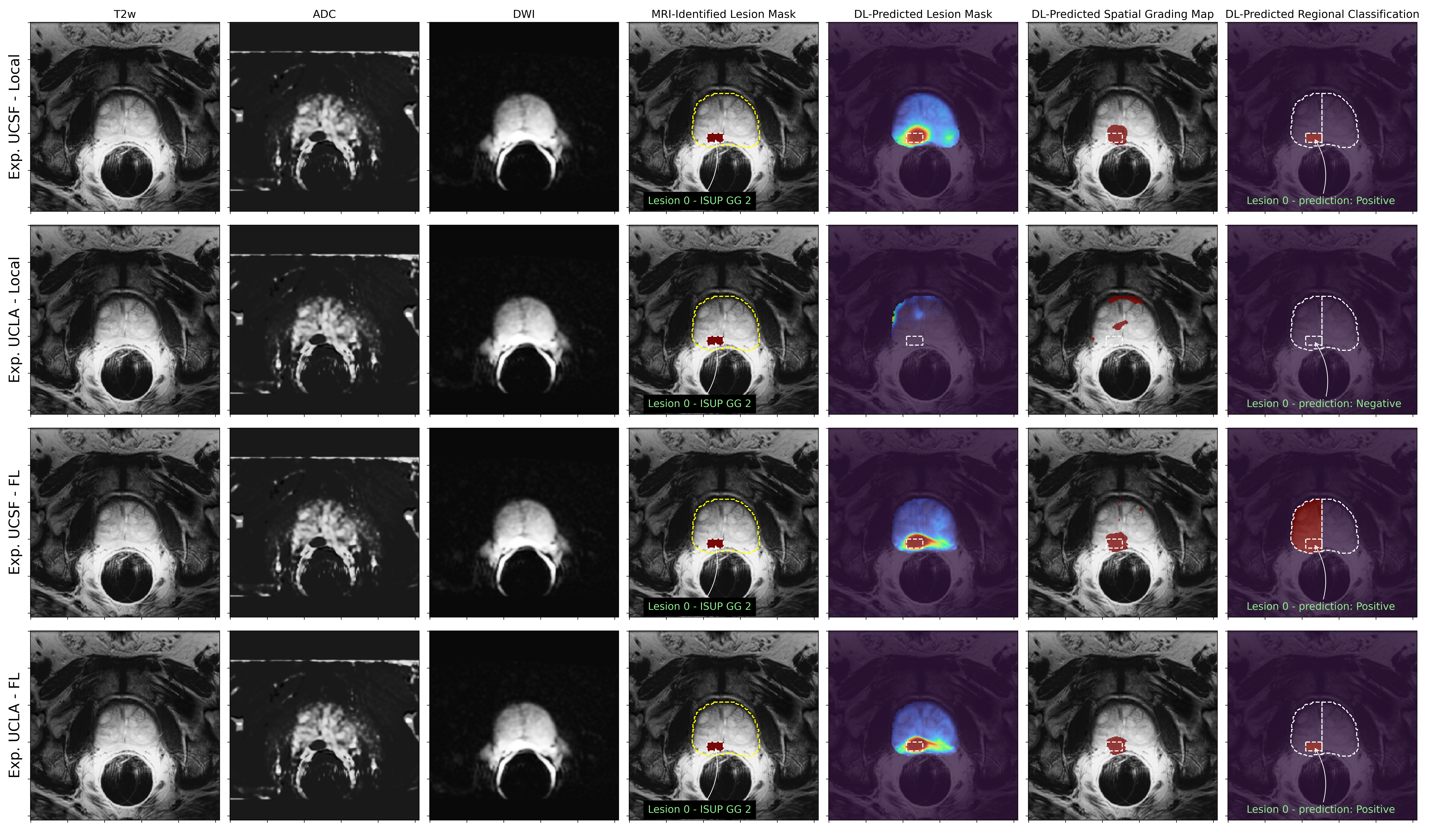}
    }
    \vfill
    \textbf{B: Evaluation of UCNet on UCSF test data, where FL \textit{improves} accuracy of UCSF-local model (Row 1 vs Rows 3-4).}\\
    {
    \centering
    \includegraphics[width=\linewidth]{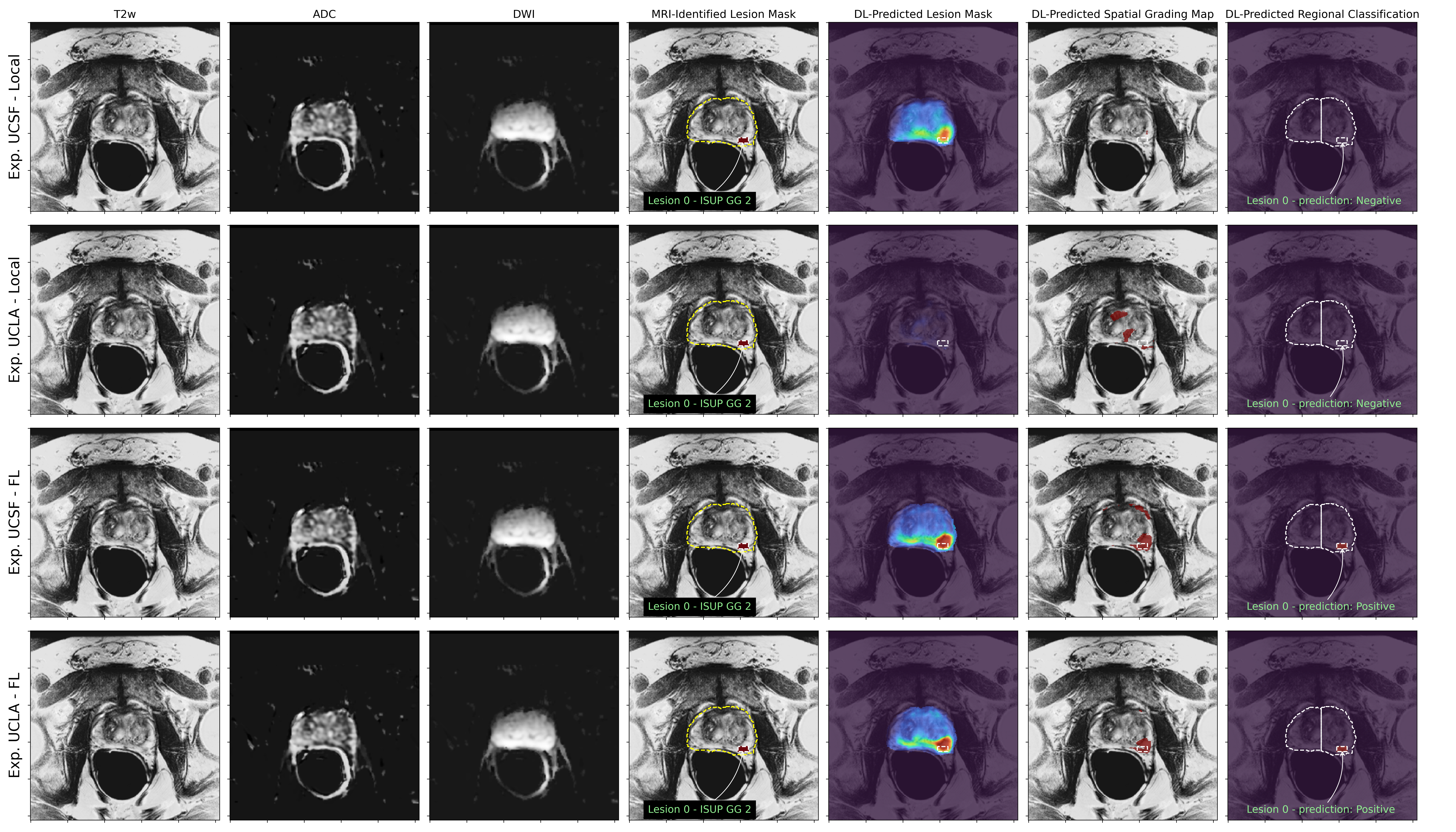}
    }
    \caption{Evaluation of UCNet on UCSF data from Exp.~ID~1111. (A)~depicts a transverse slice of an exam where UCLA's model performs poorly, but federated models (Rows 3-4) achieves the same level of accuracy as UCSF's local model. (B)~depicts a transverse slice of an exam where both local models performs poorly, but federated models perform well. Notably, for both (A-B), federated checkpoint chosen by UCLA stopping criteria (Row 3) performs better than the checkpoint chosen by UCSF (Row 4), highlighting that neither site has sufficient data to generalize well on their own.}
    \label{fig:ucsf_res}

\end{figure*}

\begin{figure*}
    \textbf{A: Evaluation on UCLA test data, where FL reclaims accuracy lost by cross-site evaluation (Row 2).}\\
    {
    \centering
    \includegraphics[width=\linewidth]{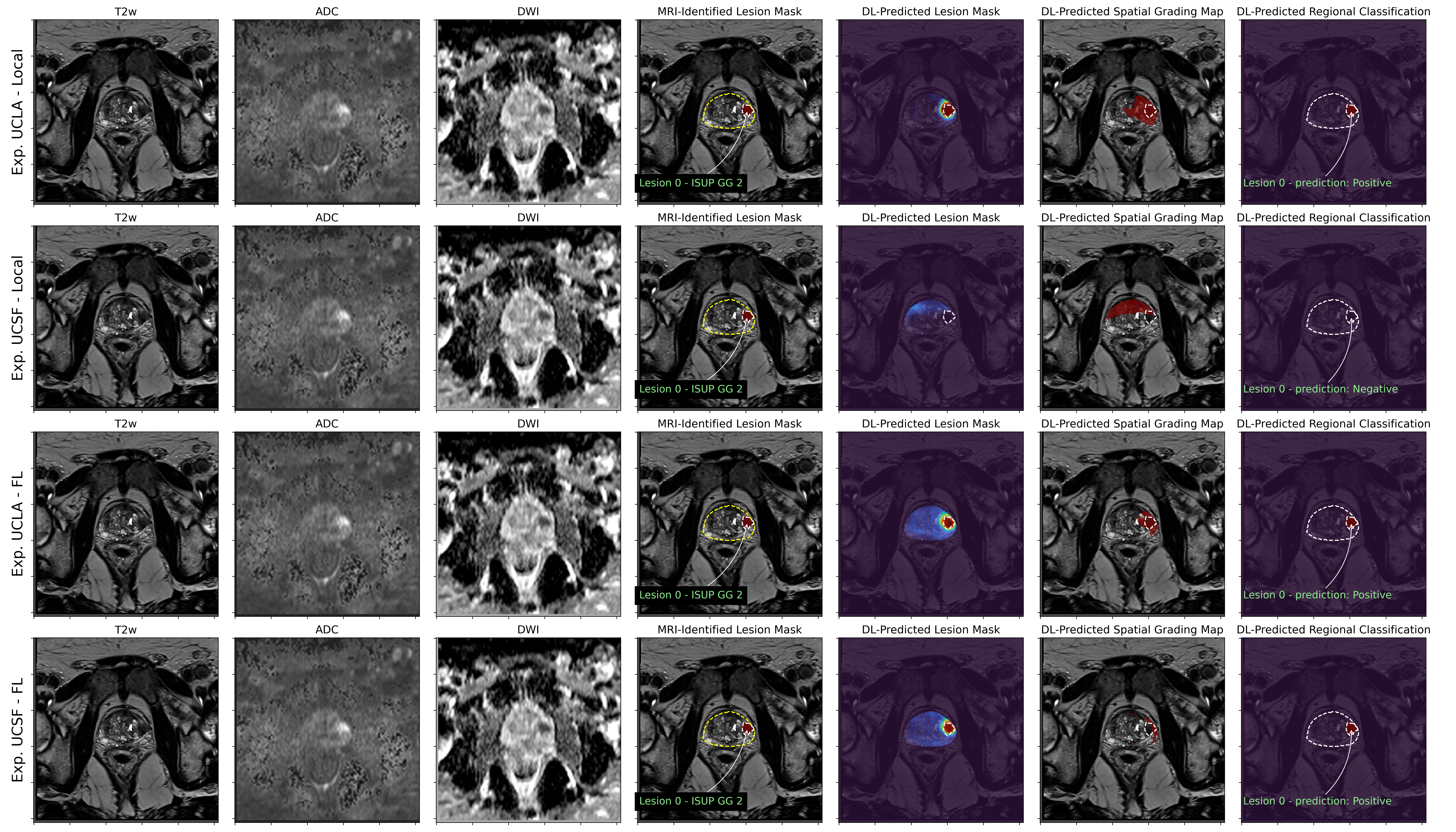}
    }
    \vfill
    \textbf{B: Evaluation on UCLA test data, where FL \textit{improves} accuracy of UCLA-local model (Row 1 vs Rows 3-4).}\\
    {
    \centering
    \includegraphics[width=\linewidth]{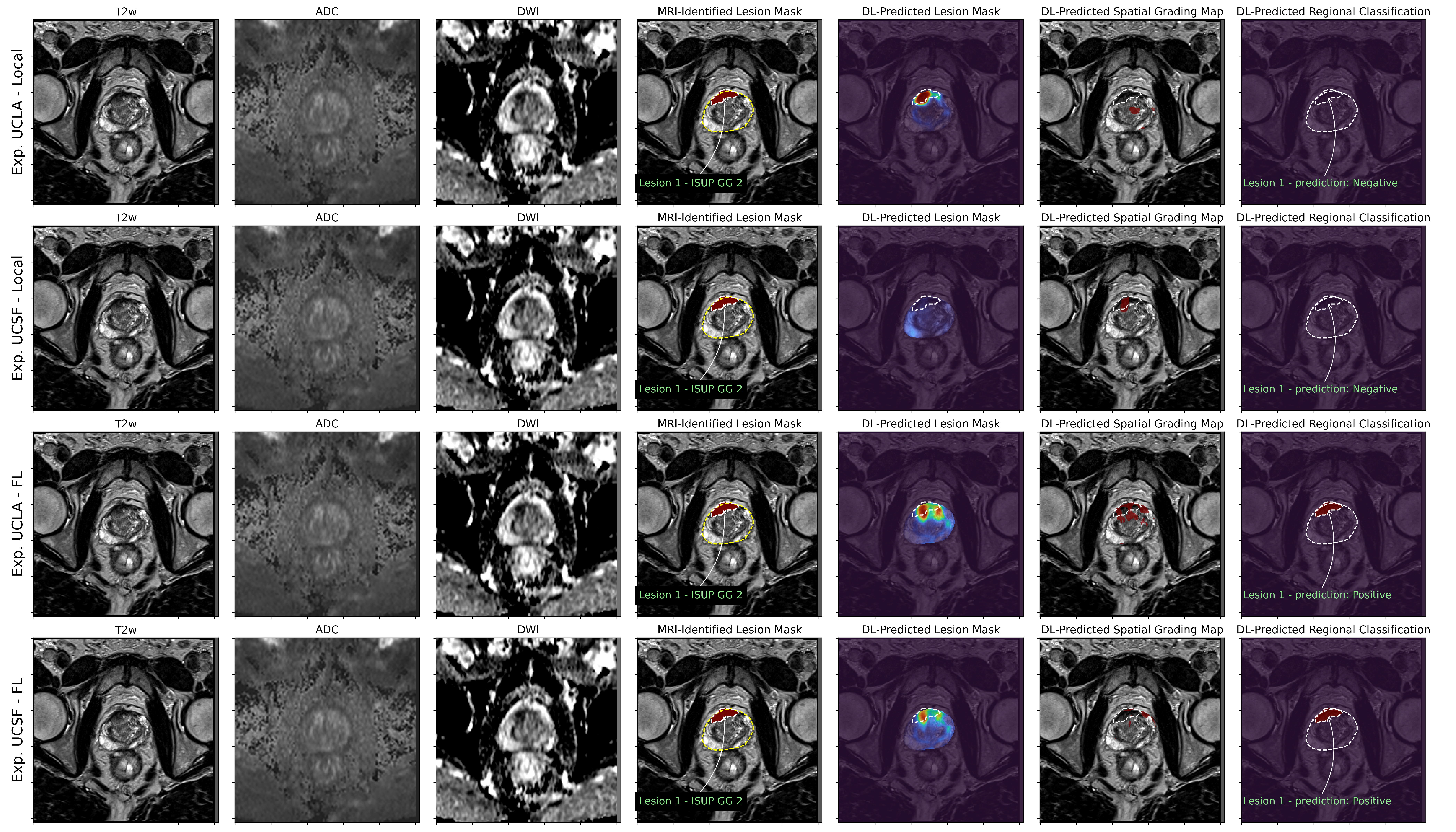}
    }
    \caption{Evaluation of UCNet on UCLA data from Exp.~ID~1111.
    (A)~depicts a transverse slice of an exam where UCSF's model performs poorly, but federated models (Rows 3-4) achieves the same level of accuracy as UCLA's local model. 
    (B)~depicts a transverse slice of an exam where both local models performs poorly, but federated models perform well. }
    \label{fig:ucla_res}
\end{figure*}

In Figures~\ref{fig:ucsf_res}~and~\ref{fig:ucla_res} we show results of local and FL models from both institutions evaluated on UCSF and UCLA test sets, respectively. As mentioned above, UCSF and UCLA local models performed well on data from the same sites, but couldn`t generalize well, which is clear in Row 2 of Figure~\ref{fig:ucsf_res}.A and Figure~\ref{fig:ucla_res}.A. The last two rows of the figures show that FL models are able to reclaim accuracy loss when cross-site evaluation is performed. 
Overall FL models evaluated on the same sites resulted in a slight accuracy drop, but still in some cases FL could improve the performance of the local model as shown in Figure~\ref{fig:ucsf_res}.B and Figure~\ref{fig:ucla_res}.B (Row 1 vs. Rows 3-4).

\section{Discussion}
In this work we addressed realistic scenario when highly heterogeneous MRI data, patient distributions, and groundtruth annotations (both within and across institutions) are available for training a deep learning algorithm. Our bespoke architecture, UCNet, was able to handle this task, and resulted in good performance in both local and federated training. The resulting models chosen on each site individually based on the local validation performance, gained the benefit of having learned from each of the private datasets without ever needing to transfer, pool, or homogenize data at a single location. Both FL checkpoints showed desirable generalization performance, and resulted in higher accuracy evaluated on the local datasets from the opposite institutions, while neither site had sufficient data to generalize well on their own.

However, the stopping criteria (for choosing checkpoints) individually on private institutional validation data should be revisited. We believe a better methodology may be to monitor global validation accuracy since the accuracy on the other institution's dataset is withheld and may give a better indicator of performance. This is evident from the deviation between within-site performance on validation and test sets, which was especially bad for UCLA with -15.4\% classification accuracy and -45\% lesion segmentation IoU.

In terms of federated learning, due to current limitations of the NVFlare framework, the server requires a small GPU to implement the central update. We believe this can be resolved in a future update to \texttt{FLtools} by performing aggregation on the central server, but pushing the weight update with all appropriate momentum terms to each client. Note that the presented \texttt{FedSGD} technique implemented in \texttt{FLComponents} achieves $O(1)$ memory complexity with respect to the number of clients, and thus is highly scalable for training with 100's of research sites.

However, one practical issue with \texttt{FedSGD} is that it has a relatively high time and communication complexity $O(n)$, since it must wait until every client responds before updating the global model \textit{at every iteration}. In this respect, \texttt{FedAvg} may be a more desirable aggregation strategy, although the training dynamics have not yet been explored with UCNet and prostate MRI. As such, one option to reduce the complexity constants for \texttt{FedSGD} may be to implement asynchronous peer-to-peer distribution of gradients and local update of weights, with the central server only acting to globally update momentum terms. We leave this as future work for \texttt{FLtools}.

\section{Conclusion}
We develop and open-source a federated learning toolkit \texttt{FLtools} that is built on NVFlare and provides extensible and reusable features to researchers with custom deep learning models and workflows, or with heterogeneous datasets that can vary in specification across clients in a federation. We successfully applied this toolkit to a custom UCNet deep learning model we developed to handle diverse prostate MRI and associated radiological and histopathology annotation, demonstrating between 9.50-14.8\% improvement in generalization performance. However, within site performance has not yet improved, indicating that the full potential of federated learning has not yet been realized for prostate MRI even with 1800+ exams. This provides motivation for inclusion of even more multi-institution prostate MRI using federated learning.

\section*{Acknowledgements}
We would like to thank Dr.~Karthik~Sarma~(UCLA) for providing insight into the radiological problem of prostate cancer classification. We would like to thank NVidia, the UCSF Center for Intelligent Imaging (CI$^2$), and Jed Chan (UCSF), for developing and supporting the network infrastructure and tooling to enable this work.

\newpage
\bibliographystyle{ieeetr}
{\small \bibliography{refs}}

\begin{thebibliography}{10}

\bibitem{sung2021global}
H.~Sung, J.~Ferlay, R.~L. Siegel, M.~Laversanne, I.~Soerjomataram, A.~Jemal,
  and F.~Bray, ``Global cancer statistics 2020: Globocan estimates of incidence
  and mortality worldwide for 36 cancers in 185 countries,'' {\em CA: a cancer
  journal for clinicians}, vol.~71, no.~3, pp.~209--249, 2021.

\bibitem{turkbey2012multiparametric}
B.~Turkbey and P.~L. Choyke, ``Multiparametric mri and prostate cancer
  diagnosis and risk stratification,'' {\em Current opinion in urology},
  vol.~22, no.~4, p.~310, 2012.

\bibitem{epstein20162014}
J.~I. Epstein, L.~Egevad, M.~B. Amin, B.~Delahunt, J.~R. Srigley, and P.~A.
  Humphrey, ``The 2014 international society of urological pathology (isup)
  consensus conference on gleason grading of prostatic carcinoma,'' {\em The
  American journal of surgical pathology}, vol.~40, no.~2, pp.~244--252, 2016.

\bibitem{westphalen2020variability}
A.~C. Westphalen, C.~E. McCulloch, J.~M. Anaokar, S.~Arora, N.~S. Barashi,
  J.~O. Barentsz, T.~K. Bathala, L.~K. Bittencourt, M.~T. Booker, V.~G.
  Braxton, {\em et~al.}, ``Variability of the positive predictive value of
  pi-rads for prostate mri across 26 centers: experience of the society of
  abdominal radiology prostate cancer disease-focused panel,'' {\em Radiology},
  vol.~296, no.~1, pp.~76--84, 2020.

\bibitem{cao2019joint}
R.~Cao, A.~M. Bajgiran, S.~A. Mirak, S.~Shakeri, X.~Zhong, D.~Enzmann,
  S.~Raman, and K.~Sung, ``Joint prostate cancer detection and gleason score
  prediction in mp-mri via focalnet,'' {\em IEEE transactions on medical
  imaging}, vol.~38, no.~11, pp.~2496--2506, 2019.

\bibitem{schelb2019classification}
P.~Schelb, S.~Kohl, J.~P. Radtke, M.~Wiesenfarth, P.~Kickingereder,
  S.~Bickelhaupt, T.~A. Kuder, A.~Stenzinger, M.~Hohenfellner, H.-P. Schlemmer,
  {\em et~al.}, ``Classification of cancer at prostate mri: deep learning
  versus clinical pi-rads assessment,'' {\em Radiology}, vol.~293, no.~3,
  pp.~607--617, 2019.

\bibitem{mehralivand2022deep}
S.~Mehralivand, D.~Yang, S.~A. Harmon, D.~Xu, Z.~Xu, H.~Roth, S.~Masoudi,
  D.~Kesani, N.~Lay, M.~J. Merino, {\em et~al.}, ``Deep learning-based
  artificial intelligence for prostate cancer detection at biparametric mri,''
  {\em Abdominal Radiology}, vol.~47, no.~4, pp.~1425--1434, 2022.

\bibitem{sarma2021federated}
K.~V. Sarma, S.~Harmon, T.~Sanford, H.~R. Roth, Z.~Xu, J.~Tetreault, D.~Xu,
  M.~G. Flores, A.~G. Raman, R.~Kulkarni, {\em et~al.}, ``Federated learning
  improves site performance in multicenter deep learning without data
  sharing,'' {\em Journal of the American Medical Informatics Association},
  vol.~28, no.~6, pp.~1259--1264, 2021.

\bibitem{antunes2022federated}
R.~S. Antunes, C.~A. da~Costa, A.~K{\"u}derle, I.~A. Yari, and B.~Eskofier,
  ``Federated learning for healthcare: Systematic review and architecture
  proposal,'' {\em ACM Transactions on Intelligent Systems and Technology
  (TIST)}, 2022.

\bibitem{nvflare}
NVIDIA, ``Nvidia federated learning application runtime environment,'' {\em
  {https://github.com/NVIDIA/NVFlare}}, 2021.

\bibitem{mcmahan2021advances}
H.~B. McMahan {\em et~al.}, ``Advances and open problems in federated
  learning,'' {\em Foundations and Trends{\textregistered} in Machine
  Learning}, vol.~14, no.~1, 2021.

\bibitem{yan2020variation}
Z.~Yan, J.~Wicaksana, Z.~Wang, X.~Yang, and K.-T. Cheng, ``Variation-aware
  federated learning with multi-source decentralized medical image data,'' {\em
  IEEE Journal of Biomedical and Health Informatics}, vol.~25, no.~7,
  pp.~2615--2628, 2020.

\bibitem{sarma2021harnessing}
K.~V. Sarma, A.~G. Raman, N.~J. Dhinagar, A.~M. Priester, S.~Harmon,
  T.~Sanford, S.~Mehralivand, B.~Turkbey, L.~S. Marks, S.~S. Raman, {\em
  et~al.}, ``Harnessing clinical annotations to improve deep learning
  performance in prostate segmentation,'' {\em Plos one}, vol.~16, no.~6,
  p.~e0253829, 2021.

\bibitem{pellicer2022deep}
O.~J. Pellicer-Valero, J.~L. Marenco~Jim{\'e}nez, V.~Gonzalez-Perez, J.~L.
  Casanova Ram{\'o}n-Borja, I.~Mart{\'\i}n~Garc{\'\i}a, M.~Barrios~Benito,
  P.~Pelechano~G{\'o}mez, J.~Rubio-Briones, M.~J. Rup{\'e}rez, and J.~D.
  Mart{\'\i}n-Guerrero, ``Deep learning for fully automatic detection,
  segmentation, and gleason grade estimation of prostate cancer in
  multiparametric magnetic resonance images,'' {\em Scientific reports},
  vol.~12, no.~1, pp.~1--13, 2022.

\bibitem{falcon2019pytorch}
W.~Falcon {\em et~al.}, ``Pytorch lightning,'' {\em {GitHub. Note:
  https://github. com/PyTorchLightning/pytorch-lightning}}, vol.~3, p.~6, 2019.

\bibitem{ronneberger2015u}
O.~Ronneberger, P.~Fischer, and T.~Brox, ``U-net: Convolutional networks for
  biomedical image segmentation,'' in {\em International Conference on Medical
  image computing and computer-assisted intervention}, pp.~234--241, Springer,
  2015.

\bibitem{rajagopal2021understanding}
A.~Rajagopal, V.~C. Madala, T.~A. Hope, and P.~Larson, ``{Understanding and
  Visualizing Generalization in UNets},'' in {\em Medical Imaging with Deep
  Learning}, pp.~665--681, PMLR, 2021.

\end{thebibliography}

\end{document}